# Causal KL: Evaluating Causal Discovery

**Rodney T. O'Donnell**  RODO@CSSE.MONASH.EDU.AU
**Kevin B. Korb**  KORB@CSSE.MONASH.EDU.AU
**Lloyd Allison**  LLOYD@CSSE.MONASH.EDU.AU
*School of Information Technology*
*Monash University*
*Clayton, Vic, Australia*

## Abstract

The two most commonly used criteria for assessing causal model discovery with artificial data are edit-distance and Kullback-Leibler divergence, measured from the true model to the learned model. Both of these metrics maximally reward the true model. However, we argue that they are both insufficiently discriminating in judging the relative merits of false models. Edit distance, for example, fails to distinguish between strong and weak probabilistic dependencies. KL divergence, on the other hand, rewards equally all statistically equivalent models, regardless of their different causal claims. We propose an augmented KL divergence, which we call Causal KL (CKL), which takes into account causal relationships which distinguish between observationally equivalent models. Results are presented for three variants of CKL, showing that Causal KL works well in practice.

## 1. THE PROBLEM

The problem which we set ourselves to solve here is: given two causal discovery algorithms $C_1$ and $C_2$, how do we decide which is the better causal learner?[1] This problem is asked and answered in almost everything published in the causal discovery literature, since almost all authors wish to demonstrate the superiority of their algorithm and attempt to do so by giving experimental results comparing one algorithm with another. When we examine the literature we find that the most common evaluative method is to: generate artificial data from a given DAG (directed acyclic graph) model $M_t(\theta)$ (with parameters $\theta$);[2] run two competing algorithms $C_1$ and $C_2$ using the artificial data, resulting in a pair of new structural models $M_1$ and $M_2$, respectively; compute an edit distance $d(\cdot, \cdot)$ from the true to the discovered models and compare them. $C_1$ is then declared superior to $C_2$ if $d(M_t, M_1) < d(M_t, M_2)$ over some sample of cases.[3] We will call any such metric $d$ a meta-metric, to distinguish it from any metric the causal learners themselves may be using.[4] The next most common technique is to substitute Kullback-Leibler divergence (Kullback and Leibler, 1951), $KL(M_t, M_i)$, for edit distance.[5] We do not believe that either of these meta-metrics is defensible for the purpose at hand.

---

1. Example causal discovery algorithms include K2 (Cooper and Herskovits, 1992), TETRAD (Spirtes et al., 2000) and CaMML (Korb and Nicholson, 2004). Note that this list includes both constraint learners and metric learners. The issue we are raising applies to all causal learners, regardless of their method. Also note that the metrics we discuss in the text for assessing learners are, and should be, distinct from any metric used by the learners themselves. Suppose, for example, that a Bayesian algorithm uses a Bayesian metric $m$ to learn its models. If the meta-metric used to assess learners were also $m$, then, of course, the Bayesian algorithm will automatically be declared the winner. While Bayesians might consider this a persuasive argument, others might wonder about the prior bias it reveals.
2. For our notation, see Appendix A.

## 1.1 EDIT DISTANCE

There are a number of possible ways to define an edit distance between DAGs. We will use the simplest.

**Definition 1 (Edit Distance)**
$ED(M_1, M_2)$ equals the minimal number of arc deletions and additions to turn $M_1$ into $M_2$.

Clearly, this is only defined for graphs over the same variables, and we will only consider such cases. We illustrate with a very simple case:

$$Lawn \leftarrow Rain \rightarrow Newspaper$$

This is intended to represent the true causal system, which is one where rainfall sometimes wets both a lawn and a newspaper. If we label that $M_1$, then, for example, the model $M_2$ = $Lawn \rightarrow Rain \rightarrow Newspaper$ will have the measure $ED(M_1, M_2) = 2$, since the first arc needs to be deleted (one operation) and then replaced with a new arc oriented oppositely (another operation).

Our objection to any such meta-metric is that it fails to attend to the strength of relationship between variables represented by different arcs. It is sensitive to the qualitative DAG, yet is being applied to models which are parameterized in ways to which edit distance is blind.

## 1.2 KULLBACK-LEIBLER DIVERGENCE

Kullback-Leibler divergence (or relative entropy), measures a kind of "distance"[6] from a true probability distribution $p$ to another distribution $q$:

**Definition 2 (KL)**

$$KL(p, q) = \sum_i p_i \log \frac{p_i}{q_i}$$

*where $i$ ranges over all possible joint instantiations of variables and by convention $0 \log 0 = 0$ and $x \log \frac{x}{0} = \infty$.[7]*

---

3. For example, see (Spirtes et al., 2000; Suzuki, 1996)
4. Of course, finding an optimal meta-metric does not fully answer our problem, since we still need to deal with such issues as getting an appropriate range of learning problems for testing the algorithms and dealing with sampling error in comparing sets of learned models. However, developing and justifying a good meta-metric is what we focus upon here.
5. For example, see (Friedman and Goldszmidt, 1996; Dash and Druzdzel, 2003).
6. We use the term divergence instead of distance as $KL(P, Q)$ is not symmetric and does not satisfy the triangle inequality.
7. KL is analogously defined (via integration) over continuous distributions.

KL reports the expected additional number of bits (assuming log base 2) needed to encode a single sample from a distribution when using another distribution for the code. This is zero when we have the true distribution and an increasingly large number as the used distribution diverges more greatly from reality. As this metric is sensitive both to DAG structure and to parameterization, the difficulty of qualitative crudeness of the edit distance metric does not apply; differences in asserting strong dependencies will count in KL commensurately more than differences in asserting weak dependencies.

So, KL might seem to be the ideal answer for our meta-metric, which is perhaps why it appears to be becoming more popular. In fact, we believe it is ideal if the learning algorithms under study are strictly probability learners. However, there is more to a causal model than a probability distribution. Networks with identical joint distributions may have drastically different causal implications. Hence, KL divergence is not an ideal meta-metric because it assesses learning algorithms on their ability to recover the true probability distribution rather than the true causal model, which is what causal discovery algorithms in fact aim for.[8] This problem is exemplified by KL's inability to distinguish between statistically equivalent models.

### 1.3 STATISTICAL EQUIVALENCE

If $M_1$ and $M_2$ are two DAGs such that for any parameter vector $\theta_1$ for $M_1$ there is a parameter vector $\theta_2$ such that $P_{M_1(\theta_1)} = P_{M_2(\theta_2)}$, and vice versa (that is, the two Bayesian networks can represent the same set of probability distributions), then $M_1$ and $M_2$ are said to be statistically equivalent. Using maximum likelihood parameterization, $M_1$ and $M_2$ will perform equally well regardless of the data (Chickering, 1995). As Verma and Pearl (1990) showed, statistical equivalence in this sense (or membership in a common *pattern*, as they called it), is equivalent to sharing the following structural properties:

1. Having the same undirected skeleton (i.e., having the same arcs, disregarding direction)

2. Having the same uncovered collisions (i.e., triples $X \rightarrow Z \leftarrow Y$, where $X$ and $Y$ are themselves not directly connected)

In consequence of this, for example, the DAGs

$$Lawn \leftarrow Rain \rightarrow Newspaper$$
$$Lawn \leftarrow Rain \leftarrow Newspaper$$
$$Lawn \rightarrow Rain \rightarrow Newspaper \qquad (P1)$$

are all equivalent (in the same pattern), while the only other triple with the same skeleton

$$Lawn \rightarrow Rain \leftarrow Newspaper \qquad (P2)$$

---

8. We note in passing that there is an ongoing discussion in the Bayesian net community about the status of causal interpretations of Bayesian nets. We are not entering that dispute here. It is not controversial that *some* Bayesian networks are causal — that is, that their arcs can be interpreted as expressing direct causal relations; we limit our interest to such networks in this paper.

is the only member of its pattern.

However, as there is no observational evidence that can distinguish between the models in P1, why penalize a learner which chooses the wrong equivalent model? KL seems to "do the right thing" here by scoring equivalent models equally. Spirtes et al. (2000), and others, even use an edit distance to the true pattern instead of to a DAG based on this logic. We will now argue against the propriety of this approach.

## 1.4 EVALUATING CAUSAL MODELS

It could be claimed that if the data cannot possibly distinguish between two models in the same pattern, then an algorithm which does distinguish between them can only be doing so on the basis of some prior bias. A meta-metric which rewards (or punishes) the learner of one model over the other is rewarding (punishing) aspects of the algorithms which have nothing to do with how well they learn from the observational evidence as such.

However, there is a substantial difference between (a) no possible observational evidence to distinguish the models and (b) no possible evidence to distinguish the models. The standard move in science to find evidence to distinguish between observationally equivalent models is to find experimental interventions which the distinct models predict will lead to distinct outcomes.

Sir Ronald Fisher, for example, (in)famously pointed out that the general assumption that the correlation between cigarette smoking and lung cancer must be explained by the former causing the latter ignored the possibility that there is a third factor, specifically a genetic factor, causing both (Fisher, 1957). If we are observing only *Smoking* and *Lung Cancer*, then we will be unable to distinguish empirically between Figure 1(a) and Figure 1(b). One solution is, of course, to also observe jointly the hidden genetic factor, in case we know how to do that. Another solution is to intervene on *Smoking* and observe the results.

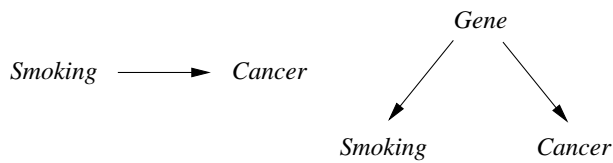

Figure 1: (a) Smoking causes lung cancer; (b) Smoking doesn't cause lung cancer.

Returning to our earlier example of (P1), interventions upon *Rain* will potentially distinguish any one of the three models from the others. Of course, we may not immediately know how to intervene upon *Rain*. But then we may either find an appropriate substitute that can still test the relevant causal hypotheses (e.g., *Sprinkler*) or be prepared to wait for technology to catch up with our scientific hypotheses. The point is that as scientists we certainly should *not* be building our present technological limitations into our theories of method. Arguably, at some point in our history there were no accessible observations to distinguish Copernican from Ptolemaic theories of our solar system; nevertheless, it was *never* the case that there was nothing at issue between the two theories.

In short, KL measures the distance between any pair of models in P1 as zero, whereas they each assert distinct causal structures which can, at least in principle, be directly tested.[9] They are directly testable by interventions upon the variables, which can be represented by augmenting the original model by adding a direct parent $X'$, "Intervene on $X$" for each original variable $X$, doubling the number of variables. We call the resultant model the (fully) augmented model (see, for example, Figure 2). As demonstrated in Korb and Nyberg (2006), any two distinct linear models, including models in a common pattern, become empirically distinguishable in the augmented space; distinct models in a common pattern are put into distinct (augmented) patterns under intervention.[10]

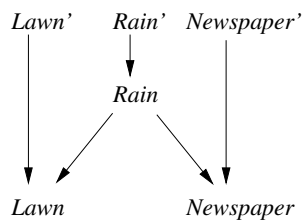

Figure 2: Augmented rain network.

What we propose, therefore, is to use KL as a meta-metric for divergence of the learned model from the true model, but with measurements conducted in the *augmented space*. We call this meta-metric Causal KL (CKL). We will develop three CKL variants subsequently.

## 2. DESIDERATA FOR A META-CRITERION

Ideally, we would now introduce a meta-meta-metric which could measure and rank our different meta-metrics, reporting in the end something like $CKL > KL > ED$ (which is something like the ideal from our biased perspective!). However, we know of no useful meta-meta-metric. The best we can do is present a list of properties which we believe are desirable for a meta-metric $m$ to have.

1. **Sensitivity:** $m$ should be more sensitive to arcs which result in stronger probabilistic dependencies between nodes.

    1a. **Subsumption of KL:** As we believe KL is an ideal meta-metric for probability distributions, $m(M_1(\theta_1), M_2(\theta_2))$ should be proportional to $KL(M_1(\theta_1), M_2(\theta_2))$ when no causal distinction exists between $M_1$ and $M_2$. That is

    $$(M_1 = M_2) \Rightarrow (KL(M_1(\theta_1), M_2(\theta_2)) \propto m(M_1(\theta_1), M_2(\theta_2))$$

---

9. A related point is that a fully connected network can represent *any* probability distribution over its variables, given the right parameterization. But unless the skeptics' view that everything is connected to everything is literally correct, such a model will rarely be right, although it will always receive a minimal KL score.
10. Related results for discrete models can be found in Nyberg and Korb (2006).

2. **Causality:** $m$ needs to be able to distinguish between DAGs within a common pattern, preferring those which represent the true causal model, other things being equal.

3. **Non-negative scalar:** $m$ should be a function into the non-negative reals; i.e., $m : Model \times Model \to [0, \infty)$

   This implies, for example, that an edit-distance metric must not report two distinct values for a DAG when there are two distinct paths of editing operations which yield that DAG. If there are multiple paths to a model, characteristics of the paths are not of interest to us, but only a "crow-flight" minimal distance between the two models.

4. **Uniqueness:** $m(M_1(\theta_1), M_2(\theta_2)) = 0$ if and only if both $M_1 = M_2$ and $\theta_1 = \theta_2$, where $M_1$ is the true DAG, $\theta_1$ is the true parameters given $M_1$, $M_2$ and $\theta_2$ are the equivalent values for the learned network.[11]

   The selection of non-negative reals with zero as the optimal value reflects the idea of $m$ being analogous to a distance metric. That zero is asserted here also to be an optimum (unique) value reflects our realist metaphysics, that there should be one and only one reality that we are dealing with at any given time.

5. **Consistency:** If $M_1 = M_2$, and $\theta_2$ is estimated from data $D$ generated by $M_1(\theta_1)$ using a consistent estimator such as maximum likelihood, then

$$\text{as } |D| \to \infty \ \ m(M_1(\theta_1), M_2(\theta_2)) \to 0$$

   We can accept that noise in the data may confuse our ability to identify the truth. Thus, an incorrect DAG may receive a lower value than the true DAG, when parameterized using a small sample. But we should like our meta-metric to reward learners which converge upon the truth when given perfect (infinite) data.

## 3. WEIGHTED EDIT DISTANCE

As a plausible alternative to $ED$ and $KL$ we attempted to create a quantitative variation, $WED$, on the standard, qualitative edit distance $ED$. Our reason for introducing $WED$ is that $ED$ itself has some nice properties, including intuitive appeal. $ED$ is also directly responsive to the causal DAG structure, penalizing variants within the original pattern that get the causal story wrong. The only complaint we know of is its inability to weight lightly arcs which carry only weak probabilistic dependencies or weight heavily those which carry strong dependencies; all edit operations are given equal weight. A weighted $ED$ metric might keep the good properties while addressing $ED$'s failings.

---

11. To be exact, this requires there be no strictly spurious arcs; that is, the models under test need to be minimal Markov models. We assume this. Note that no *sensible* causal discovery algorithm adds spurious arcs. (And also note that this issue is distinct from the question of faithfulness: true causal models may well be unfaithful, implying that a simpler model can represent the true probability distribution. However, such models are still minimal Markov models.)

First, we describe a weighted edit distance for path models, since it is easiest for that case; however, we have done experimental work only for discrete models, which we introduce subsequently.

As a first attempt a direct and simple means of weighing arc deletions and additions is considered. In the case of standardized linear models (path models) the most obvious weight to use for $X \xrightarrow{p_{YX}} Y$ would be the square of the path coefficient, $p_{YX}^2$, which is just the amount of variance in the child node which is directly due to the parent node. The discrete analogue to this is mutual information, the amount of information about the child variable which is communicated by the parent variable — or, alternatively, the expected change in optimal code length for the child given the parent. Thus, we have two versions of $WED$, one for linear and one for discrete models.

**Definition 3 ($WED_P$)**
$$WED_P(M_1, M_2) = \min_{\Phi} \sum_{i,j} p_{ji}^2$$

where $M_1$ and $M_2$ are path models.

**Definition 4 ($WED_D$)**
$$WED_D(M_1, M_2) = \min_{\Phi} \sum_{i,j} I(X_i, X_j)$$

where $M_1$ and $M_2$ are discrete Bayesian networks, $I(\cdot, \cdot)$ is mutual information [12]

In each case we require the meta-metric to select the path $\Phi$ of edits that minimizes the result (i.e., this is a "crow-flight" distance).

However there are problems with $WED_D$, unlike the linear models used in $WED_P$, $WED_D$ has to deal with non additive dependencies. This is exemplified by the case where $C$ is a child of $A$ and $B$ with the relationship $C = A\ XOR\ B$ where $I(A, C) = I(B, C) = 0$.

Revisiting our desiderata, as $WED_P$ and $WED_D$ only take account of $\theta_1$, and not $\theta_2$, the parameters of the model being tested, both fail our Uniqueness Desideratum. Though generally more sensitive than unweighted ED neither $WED_P$ or $WED_D$ come close to satisfying Desideratum 1a.

We were unable to devise a pairwise WED function which satisfactorily fulfilled our desiderata. This inspired us to broaden our search to an "Edit Distance" over a variable's parent set instead of the traditional pairwise approach.[13]

In the next section we examine the new causal KL approach, then return to discuss a parentset weighted ED in Section 4.4.

---

12. We note that mutual information does not take into account "pass through" dependencies in the same way the $WED_P$ does. "$A \to B \to C, A \to C$" will have a larger than required weight on $A \to C$.
13. This new meta-metric is not a true Edit Distance, but as it was inspired by ED we retain the name.

# 4. CAUSAL KULLBACK-LEIBLER DIVERGENCE

To recapitulate, we desire a meta-metric for assessing causal models. We believe KL divergence is ideal in its place, i.e., for comparing probability distributions, but when used conventionally it ignores causality. To overcome this limitation CKL calculates a KL distance between networks augmented by a set of intervention variables $X'$. Each variable $X'_i$ in $X'$ acts as an intervention forcing a new distribution over $X_i$.

This is analogous to a real world scenario in which we are able to intervene on any variable in a system; e.g., we can force a subject to smoke or give them lung cancer directly. This type of intervention is done every day by scientists to reveal cause and effect. In all real world scenarios experiments consume resources, including time and money, so we wish to maximize our information gain and minimize expenses from a finite number of experiments. This often takes the form of intervening on a single variable, such as forcing half of the subjects to smoke, and observing the remaining variables.

For this idealized CKL, on the other hand, resources are (effectively) unlimited. Instead of running a finite set of experiments, is it possible to run *every* experiment, weighting the results by a distribution over intervention variables. A combination of experiment types, each intervening on a different variable or sets of variables, may be required to distinguish between a given set of models.

These augmentations allows discrimination between statistically equivalent original models by throwing all (distinct) models into different equivalence classes in the augmentation space. In particular, wherever two models differ on the orientation of an arc, say one asserts $A \to B$ and the other $B \to A$, in the augmented space they will have distinct collisions and so be in distinct patterns: the first having $A \to B \leftarrow B'$ and the second $B \to A \leftarrow A'$ (Korb and Nyberg, 2006).

Formally, we can define CKL as:

**Definition 5 (CKL)**

$$CKL(P_1, P_2) = \sum_{\vec{x}'} \sum_{\vec{x}} P'_1(\vec{x}', \vec{x}) \log \frac{P'_1(\vec{x}', \vec{x})}{P'_2(\vec{x}', \vec{x})}$$

*where $\vec{x}'$ and $\vec{x}$ range over instantiations of intervention and original variables and $P'_1$ and $P'_2$ extend the probability distributions $P_1$ and $P_2$, respectively, to the fully augmented space.*

This definition is ambiguous, since there are (infinitely) many ways of extending a probability distribution $P(\mathbf{X})$ to $P'(\mathbf{X}, \mathbf{X}')$ — that is, there are any number of ways of conducting experimental interventions upon the original variables. In addition, there are many possible types of intervention on a variable (see, e.g., Korb et al., 2004).

We consider perfect interventions on one or more variables. What we use for our $CKL$ variants here are interventions $X'_i$ which set the value of $X_i$ and only that value. The number of states of $X'_i$ is $|X_i| + 1$; that is, there is one state for each state of the target (intervened upon) variable and one additional state specifying that no intervention has been applied.

Before proceeding some notation must be introduced. First, we use $= \oslash$ for the "non intervention state". We define "the intervention set" to be the set of variables in $\mathbf{X}'$ actively intervening upon **X**, that is all $X'_i$ where $X'_i \neq \oslash$. The "intervention values" are the values the intervention set takes. The intervention set and intervention values together define a distribution over the intervention space $\mathbf{X}'$, with the first specifying what should be intervened upon and the second the value given under intervention.

Three variants of $CKL$ are now introduced. We believe these are plausible interpretations (specializations) of the $CKL$ meta-metric. The first two, $CKL_1$ and $CKL_2$ satisfy most of our desiderata, with the third variant, $CKL_3$, satisfying all the desiderata and having an additional justification based on weighted Edit Distance. The metastatic cancer example shown in Figure 3 is used to illustrate these meta-metrics.[14]

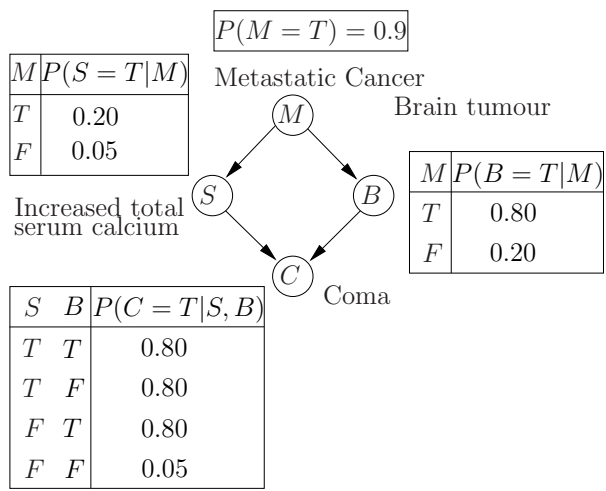

Figure 3: Original (unaugmented) metastatic network.

### 4.1 $CKL_1$: UNIFORM DISTRIBUTION

$CKL_1$ applies a uniform distribution over both the intervention set and the intervention values. That is, each $X'_i$ has a 0.5 probability of taking the "no intervention" state; the remaining probability is divided equally among the states of $X_i$, such that if an intervention occurs, then $X_i$ itself is uniformly distributed. More formally:

- $\forall i \ P'(X'_i = \oslash) = 0.5$

- $\forall ij \ P'(X'_i = x_j) = \frac{1}{2|X_i|}$ where $x_j \neq \oslash$.

- $\forall ij \ P'(X_i = x_j | X'_i = x_j) = 1$ where $x_j \neq \oslash$.

- $\forall ij \ P'(X_i = x_j | X'_i = \oslash, \pi'(X_i)) = P(X_i = x_j | \pi(X_i))$

---

14. The prospects for interventions to produce or eliminate cancers in the way described above are not to the point, as we have argued. We have selected this example simply because it is both simple and widely used in the Bayesian net literature.

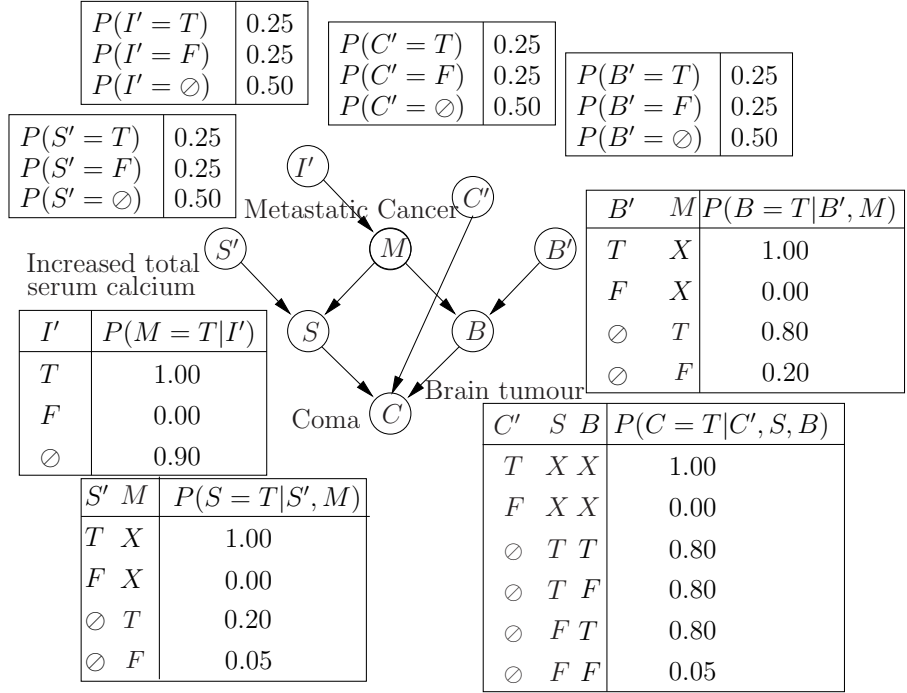

Figure 4: Metastatic network augmented with uniform distribution.

Starting from Figure 3 the $CKL_1$ augmentation gives us Figure 4. By construction, the augmented network yields precisely the original distribution, when we condition upon all intervention variables being in the non-intervention states (i.e., $\forall i : X'_i = \oslash$). However, the interventions alter the marginal (and other conditional) probabilities over the original variables. For example, pre-augmentation $P(M = T) = 0.9$, whereas post-augmentation we have $P'(M = T) = 0.7$.

$CKL_1$ satisfies most of our desiderata, however it still has some problems. $CKL_1$ satisfies Sensitivity but is not equivalent to $KL$ when $M_1 = M_2$ and as such doesn't satisfy Desideratum 1a. Clearly, $CKL_1$ also satisfies Desiderata 2 and 3. With the minor caveat that $CKL$ cannot distinguish between a zero-weighted arc and no arc (which are effectively the same thing), $CKL_1$ satisfied Uniqueness.

While the uniform choice of intervention set seems plausible, the uniform distribution over intervention values does not. Consider the network Gender $\rightarrow$ Hormone $\leftarrow$ Pregnant in which gender and being pregnant effect some hormone level. The $CKL_1$ score will be effected by the true and learned model's predictions about the hormone level of pregnant males when we intervene to set Gender = Male and Pregnant = True, even though it may be physically impossible to augment the true model in the way required by $CKL_1$, which seems unreasonable. In less extreme examples $CKL_1$ will overweight interventions on unlikely combinations of events and underweight likely combinations. A plausible solution is to use the true network to define a distribution over the augmented space.

## 4.2 $CKL_2$: TRUE MODEL DISTRIBUTION

$CKL_2$ adopts no fixed prior distribution over intervention values. $CKL_1$'s uniform distribution over the intervention set is retained; however, the remaining probability mass is chosen so that it yields a distribution over the original variables that is identical to that of the original model over those variables.

$CKL_2$ is best explained using a three tier process, as shown in Figure 5. The first tier $(T_M, T_S, T_B, T_C)$ is a copy of the true network; this is used to compute the distribution over intervention node states. The second tier $(M', S', B', C')$ of intervention variables is used to choose the intervention set. If a variable is in the intervention set it "passes through" values from the first tier, otherwise is takes the state $\oslash$ and its child retains its original distribution. The third tier $(M, S, B, C)$ contains the original network to be augmented (be it true or learned).

On the second tier, each node has a 50/50 chance of making no intervention or else of passing through the value from the previous tier. This gives an equal probability of each possible set of interventions, and a probability based on the true distribution for values of variables in the intervention set.

The first two tiers could be collapsed into one without affecting the joint distribution over $\mathbf{X}'$ or the $CKL_2$ score. The third tier of $CKL_2$ is parameterized as per the second tier of $CKL_1$.

Formally:

- $\forall i\ P'(X_i' = \oslash) = 0.5$

- $P'(\vec{a}') = P(\vec{a})$ where $\mathbf{A}' \subseteq \mathbf{X}'$ is the intervention set and $\vec{a}$ any instantiation of $\mathbf{A}$.

- $\forall ij\ P'(X_i = x_j | X_i' = x_j) = 1$ where $x_j \neq \oslash$.

- $\forall ij\ P'(X_i = x_j | X_i' = \oslash, \pi'(X_i)) = P(X_i = x_j | \pi(X_i))$

In most tested cases $CKL_2$ works well. Unlike $CKL_1$ the marginal distributions in the augmented network match those of the unaugmented net. Common variable combinations are more likely to be intervened upon as desired. It can even be shown that this metric satisfies Desideratum 1a when $M_1 = M_2$ and $M_1$ is a tree like network in which no node has more than one parent. However, this is not true in general.

Unlike $CKL_1$, $CKL_2$ will never intervene to set both Gender=Male and Pregnant=True simultaneously, as this combination never occurs in the true network. However $CKL_2$ can intervene to set Pregnant=True while observing Gender=Male. So, $CKL_2$ still suffers from this kind of problem of modeling impossible situations, although to a lesser degree. In short, $CKL_2$ passes and fails the same desiderata as $CKL_1$, with the added bonus that its sensitivity is closer to that of $KL$, and noting that it satisfies Desideratum 1a on the limited set of tree-based networks.

Next we introduce $CKL_3$ which we believe is an ideal form of CKL that satisfies all of our desiderata.

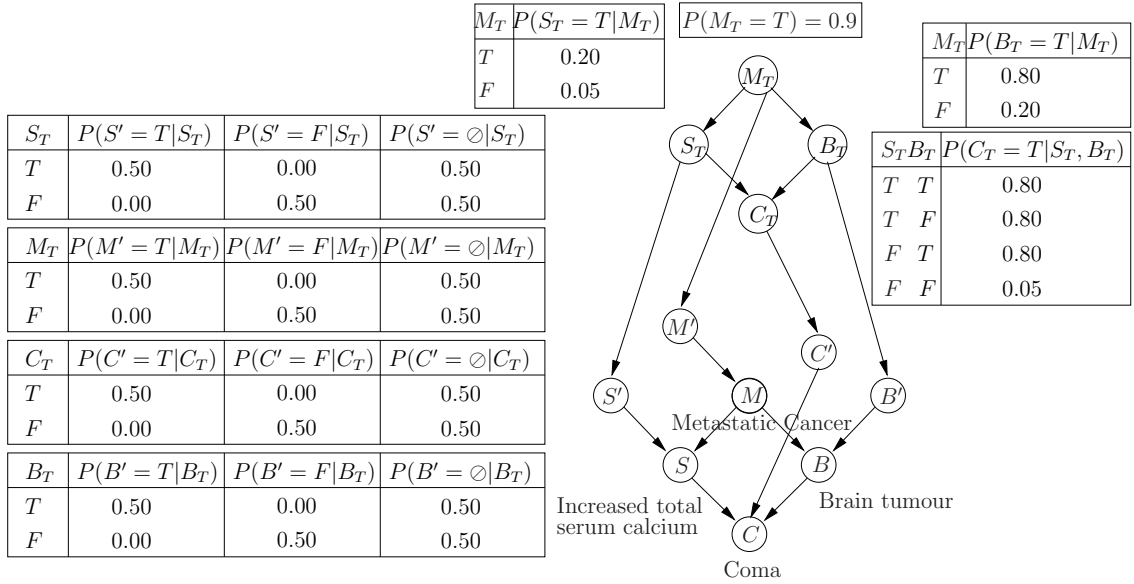

Figure 5: Metastatic network augmented with original model's distribution.

### 4.3 $CKL_3$: OPTIMAL DISTRIBUTION

Our choice of prior distributions for $CKL_1$ and $CKL_2$, while convenient, was somewhat arbitrary, although both satisfy most of our desiderata. We now develop a more principled approach to find an optimal augmentation for assessing causal models.

It was mentioned in Section 1.2 (and Desideratum 1a) that we believe KL to be ideal for assessing a joint probability distribution. If we have two networks $M_1$ and $M_2$ parameterized by $\theta_1$ and $\theta_2$ respectively and $M_1 = M_2$, the "gold standard" measure of divergence is $\propto KL(M_1(\theta_1), M_2(\theta_2))$. That is, when two models agree on the causal structure of a system, all that distinguishes them is a probability distribution. As such KL is the ideal measure of divergence between the two systems.

With this in mind, our ideal CKL meta-metric should have the property

$$KL(M_1(\theta_1), M_2(\theta_2)) = CKL(M_1(\theta_1), M_2(\theta_2)) \times c + d$$

when $M_1 = M_2$.

We have shown there is only one possible type of augmentation that both satisfies this restriction and maximally discriminates between causally distinct models (see Appendix B). This augmentation is similar to that of $CKL_2$, but instead of using a uniform distribution over all intervention sets, it uses a uniform distribution over all intervention sets containing exactly one non-intervention value. That is, a single variable is observed while the remaining $k-1$ variables are intervened upon. As with $CKL_2$ the distributions over intervention values is taken from the prior network.[15]

This may seem counterintuitive, and somewhat backwards, especially when comparing $CKL_3$ to real world practice where single interventions are most commonly used. This discrepancy may be due to economic and physical limitations in real applications, which are not a factor when studying artificial models. We could argue that if costs (or ethical or technological barriers to intervention) were removed, a scientist would be better able to discriminate between competing models of the world using all-but-one interventions rather than single variable interventions.

---
15. This is modelled by adding a "selector" node to Figure 5 as shown in Figure 6.

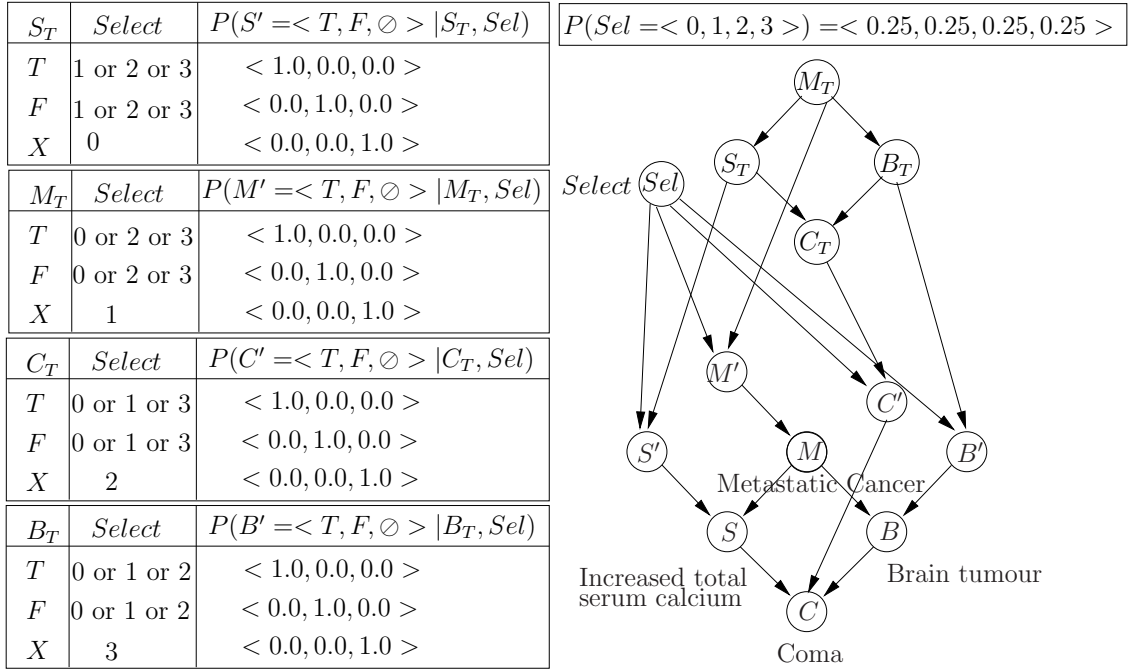

Figure 6: Metastatic network augmented with original model's distribution and uniform prior over all single non-intervention intervention sets.

Defining $CKL_3$ formally:

- $\forall i\ P'(X'_i = \oslash) = 1/c$
  Each node $X'_i$ is set to $\oslash$ an equal portion of time.

- $\forall ij$ if $X'_i = \oslash$ and $i \neq j$ then $x'_j \neq \oslash$
  Only a single node $X'_i$ may take the value $\oslash$ at any time.

- $P'(\vec{a}') = P(\vec{a})$ where $\mathbf{A}' \subseteq \mathbf{X}'$ is the intervention set and $\vec{a}$ any instantiation of $\mathbf{A}$.
  The distribution over interventions $P'(\mathbf{X}')$ is equivalent to the original unaugmented model $P(\mathbf{X})$.

- $\forall ij\ P'(X_i = x_j | X'_i = x_j) = 1$ where $x_j \neq \oslash$
  All interventions are perfect interventions.

- $\forall ij\ P'(X_i = x_j | X'_i = \oslash, \pi'(X_i)) = P(X_i = x_j | \pi(X_i))$
  The distribution over $X_i$ is identical in $P$ and $P'$ when $X'_i = \oslash$.

As desired, $CKL_3$ fulfills Desideratum 1a. When $M_1 = M_2$ $KL = CKL_3 \times c$, where $c$ is the number of nodes in **X**. Returning to the pregnant male example, like $CKL_2$, $CKL_3$ will never intervene to set both Gender = Male and Pregnant = True simultaneously. Unlike $CKL_2$ there is no way to set Pregnant = True while observing Gender = Male.

$CKL_3$ fulfills all our desiderata, though it is more complex and less intuitivethan other meta-metrics. In the next section we describe a function inspired by weighted edit distance that gives identical results to $CKL_3 \times c$ but is less complex and more intuitive.

### 4.4 Parentset Weighted Edit Distance

While searching for a weighted ED metric to compare with CKL,[16] a satisfactory WED meta-metric which only used pairwise relationships could not be found. Instead we adopted a parentset-by-parentset approach. For each node $X_i$ the sum of weighted KL distances is taken between $P_1(X_i|\pi_1(X_i))$ and $P_2(X_i|\pi_2(X_i))$ where $\pi_1(X_i)$ and $\pi_2(X_i)$ are the set of $X_i$'s parents in networks $M_1$ and $M_2$, respectively.

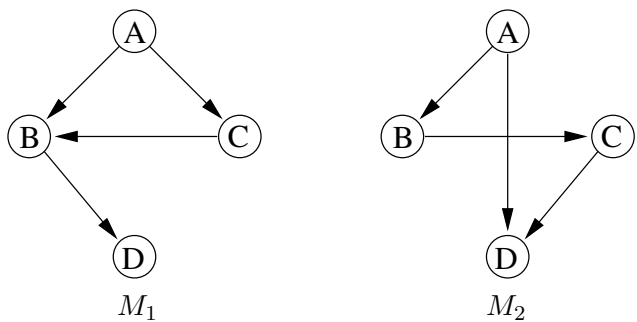

Figure 7: Two DAGs, $M_1$ and $M_2$, as used in Weighted ED example

For $M_1$ and $M_2$ in Figure 7 our Weighted ED becomes:

$$
\begin{aligned}
WED_3 &= \sum_{a} P_1(a) \times \log \frac{P_1(a)}{P_2(a)} \\
&+ \sum_{\langle a,b,c \rangle} P_1(a,c) \times P_1(b|a,c) \times \log \frac{P_1(b|a,c)}{P_2(b|a)} \\
&+ \sum_{\langle a,b,c \rangle} P_1(a,b) \times P_1(c|a) \times \log \frac{P_1(c|a)}{P_2(c|b)} \\
&+ \sum_{\langle a,b,c,d \rangle} P_1(a,b,c) \times P_1(d|b) \times \log \frac{P_1(d|b)}{P_2(d|a,c)}
\end{aligned}
$$

---

16. More specifically, to compare with $CKL_2$, which then led to the discovery of $CKL_3$.

or more generally:

$$WED_3 = \sum_i \sum_{\vec{y}_{1i},\vec{y}_{2i}} P_1(\vec{y}_{1i},\vec{y}_{2i}) P_1(x_i|\vec{y}_{1i}) \ln \frac{P_1(x_i|\vec{y}_{1i})}{P_2(x_i|\vec{y}_{2i})} \qquad (1)$$

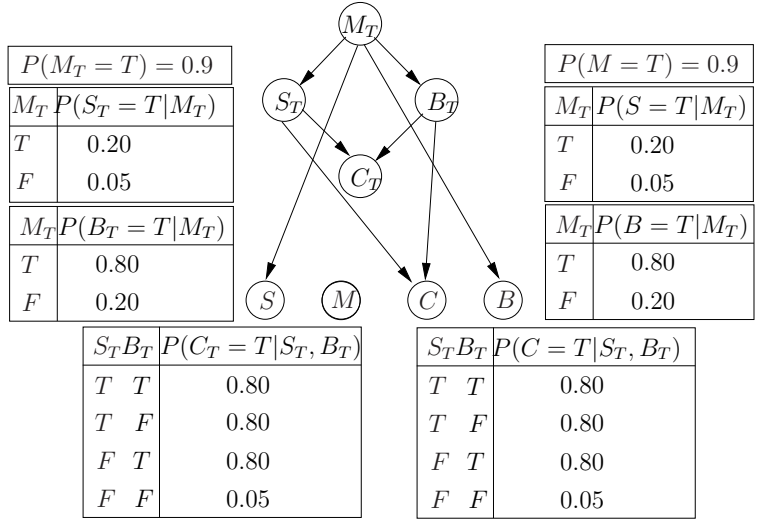

Figure 8: Metastatic network augmented for $WED_3$.

Surprisingly, this function is equivalent to $CKL_3 \times k$, as is straightforward to prove. An augmented $CKL_3$ network (Figure 6) may be converted to an augmented $WED_3$ network (Figure 8) by the following steps:

1. For each value of "select" in the $CKL_3$ network, create a new copy of the network with the value of select fixed. This introduces the term of factor $k$.

2. With select fixed, all arcs from the top layer (where intervention values are chosen) to the middle layer (where intervention values are passed through) become deterministic. Likewise all links from the middle layer to the unselected nodes in the bottom layer become deterministic. As such it is possible to remove all these deterministic arcs and nodes (as they have zero effect on the KL calculation) and hook the parents of the selected node, $X_i$, directly to its parent equivalents in the top layer. So, if in the metastatic network $C$ was selected, instead of the deterministic chain $T_s \to S' \to S \to C$, the direct connection $T_s \to C$ would give equivalent results.

3. With the deterministic relationships removed, each of the networks representing a different selection node would contain the true network and a single selection node. These networks can be combined into a single network, as in Figure 8, for the purposes of calculating KL divergence.

The reformulation of $CKL_3$ in Equation 1 gives both a simplified method of calculating $CKL$ and an alternative motivation for the metric, namely as a natural way of incorporating the strength of probabilistic dependency into a measure based upon edit distance. Of course, as $WED_3$ is equivalent to $CKL_3$, it also satisfies all our desiderata.

In this section we have have developed three unique variants of our $CKL$ metric. A fourth metric has been developed and shown to be equivalent to our best $CKL$ while also being easier to work with. We claim that the CKL metrics fulfill our desiderata; in the next section we show results confirming these claims.

## 5. THEORETICAL RESULTS

As no plausible meta-meta-metric is available, we consider the performance of our seven meta-metrics in the light of our desiderata. First we consider what should be true theoretically, then we use the metastatic model (Figure 3) as a concrete example to test these claims.

Table 1 lists how each meta-metric performs relative to our desiderata.

| meta-metric | Sensitive D1 | D1a | Causal D2 | Scalar D3 | Unique D4 | Consistent D5 |
|---|---|---|---|---|---|---|
| $ED$ | N | N | Y | Y | N | Y |
| $WED_P$ | Y | N | Y | Y | N | Y |
| $WED_D$ | Y | N | Y | Y | N | Y |
| $KL$ | Y | Y | N | Y | N | Y |
| $CKL_1$ | Y | N | Y | Y | Y* | Y* |
| $CKL_2$ | Y | N | Y | Y | Y* | Y* |
| $CKL_3/WED_3$ | Y | Y | Y | Y | Y | Y |

Table 1: Desiderata satisfaction table. *Assuming strictly positive joint probability distribution (JPD).

1. **Sensitivity:** KL (and so $CKL$), path coefficients and mutual information are all sensitive to the strength of the direct probabilistic dependence carried by causal arcs. Only KL and $CKL_3$ satisfy the stricter Desideratum 1a. In the pathological case of an $XOR$ based network $WED_D$ can be less sensitive than $ED$.

2. **Causality:** Edit distance is directly sensitive to causal orientation, whether within or between patterns. $CKL$ is indirectly sensitive to the same, by way of introducing new colliders to break up patterns. KL cannot distinguish DAGs within patterns.

3. **Non-negative Scalar:** By construction, all meta-metrics map into $[0, \infty)$.

4. **Uniqueness:** By construction, all meta-metrics are minimal at zero for the case where $M_1 = M_2$ and $\theta_1 = \theta_2$. In particular, for $KL$ metrics summands for identical terms are zero; for weighted edit distance of identical substructures no addition or deletion operators are needed and so there will be no corresponding summands.

   As ED metrics do not take account of parameters, they are not uniquely minimized by the true $\theta$. KL gives a zero score to any model in the true equivalence class. As such ED and KL are not unique metrics. CKL is unique as it gives a uniquely zero score to the true $(M, \theta)$ pair.

| Name | Mutation | Graph | Edit Distance | $WED_2$ |
|---|---|---|---|---|
| true | no Mutation | | 0 | 0 |
| tweak.weak | $P(C=T\|S=T, B=F) = 0.75$ | | 0 | 0 |
| tweak.strong | $P(C=T\|S=F, B=T) = 0.75$ | | 0 | 0 |
| add.weak | $S \to B$ added | | 1 | $I(S,B)$ |
| add.strong | $S \to B$ and $M \to C$ added | | 2 | $I(S,B) + I(M,C)$ |
| del.weak | $M \to S$ removed | | 1 | $I(S,B)$ |
| del.strong | $M \to B$ removed | | 1 | $I(M,B)$ |
| rev.in.weak | $M \to S$ reversed | | 2 | $2 \times I(M,S)$ |
| rev.in.strong | $M \to B$ reversed | | 2 | $2 \times I(M,B)$ |
| rev.out.weak | $S \to C$ reversed | | 2 | $2 \times I(S,C)$ |
| rev.out.strong | $B \to C$ reversed | | 2 | $2 \times I(B,C)$ |

Table 2: Mutations to metastatic model

5. **Consistency:** Assuming $P_1(\mathbf{X})$ is strictly positive and a consistent parameterization algorithm is used (e.g., maximum likelihood estimation), the disturbances caused by noise will disappear in the limit. The true structural model under these conditions will (in the limit) receive a score of 0 on all meta-metrics.

   When $P_1(\mathbf{X})$ is not strictly positive, $CKL_1$ and $CKL_2$ do not approach 0 in the limit and may be undefined due to conditioning on impossible combinations of states.

### 5.1 Practical Results: Metastatic Model

Here we describe experiments on the Metastatic model to test the meta-metrics in a "real world" scenario. To simulate the results returned by different BN learners, the true metastatic model was mutated by adding/deleting/reversing an arc, or tweaking a parameter, as shown in Table 2. The distance from the true network to the mutated network is then measured using each meta-metric. The resulting networks are parameterized using either finite data (1000 samples) or the true joint probability distribution (JPD; effectively infinite data). To reduce problems with noisy datasets, results presented are the means and standard deviations over 1000 tests. (Obviously, there is no need for multiple tests when parameterizing directly from the JPD.)

Table 2 shows all mutations to the metastatic dataset. Mutations labeled "del" are missing an arc present in the true model, "add" have an extra arc, "rev.in" and "rev.out" have a reversed arc and are inside or outside the true equivalence class respectively. "Tweaked" models have the correct structure, but with a parameter changed. Each mutation has a "strong" and a "weak" version. We expect "Strong" mutations to have a larger effect on the JPD than their weak counterpart, though not necessarily more than a different style of weak mutation.

| Mutation | $KL$ | $CKL_1 \times 2$ | $CKL_2 \times 2$ | $CKL_3 \times c$ |
|---|---|---|---|---|
| true | 0.0 | 0.0 | 0.0 | 0.0 |
| tweak.weak | 0.0003 | 0.0010 | 0.0003 | 0.0003 |
| tweak.strong | 0.0042 | 0.0026 | 0.0042 | 0.0042 |
| add.weak | 0.0 | 0.0 | 0.0 | 0.0 |
| add.strong | 0.0 | 0.0 | 0.0 | 0.0 |
| del.weak | 0.0087 | 0.0246 | 0.0087 | 0.0087 |
| del.strong | 0.0727 | 0.1982 | 0.0727 | 0.0727 |
| rev.in.weak | 0.0 | 0.0357 | 0.0105 | 0.0210 |
| rev.in.strong | 0.0 | 0.2080 | 0.0749 | 0.1499 |
| rev.out.weak | 0.0411 | 0.1569 | 0.0561 | 0.0655 |
| rev.out.strong | 0.0739 | 0.3115 | 0.2191 | 0.3560 |

Table 3: Effects of mutations on meta-metrics given infinite data. CKL values scaled by $P(X_i = \oslash)$ for direct comparison to KL.

| Mutation | $KL$ | $CKL_1 \times 2$ | $CKL_2 \times 2$ | $CKL_3 \times c$ |
|---|---|---|---|---|
| true | $0.0045 \pm 0.0022$ | $0.0077 \pm 0.0047$ | $0.0045 \pm 0.0022$ | $0.0045 \pm 0.0022$ |
| add.weak | $0.0054 \pm 0.0024$ | $0.0139 \pm 0.0110$ | $0.0056 \pm 0.0026$ | $0.0054 \pm 0.0024$ |
| add.strong | $0.0069 \pm 0.0028$ | $0.0247 \pm 0.0150$ | $0.0078 \pm 0.0031$ | $0.0069 \pm 0.0028$ |
| del.weak | $0.0127 \pm 0.0021$ | $0.0309 \pm 0.0046$ | $0.0127 \pm 0.0021$ | $0.0127 \pm 0.0021$ |
| del.strong | $0.0767 \pm 0.0021$ | $0.2045 \pm 0.0097$ | $0.0767 \pm 0.0021$ | $0.0767 \pm 0.0021$ |
| rev.in.weak | $0.0045 \pm 0.0022$ | $0.0428 \pm 0.0103$ | $0.0150 \pm 0.0042$ | $0.0254 \pm 0.0072$ |
| rev.in.strong | $0.0045 \pm 0.0022$ | $0.2149 \pm 0.0128$ | $0.0794 \pm 0.0076$ | $0.1542 \pm 0.0146$ |
| rev.out.weak | $0.0454 \pm 0.0022$ | $0.1644 \pm 0.0117$ | $0.0605 \pm 0.0037$ | $0.0694 \pm 0.0054$ |
| rev.out.strong | $0.0784 \pm 0.0024$ | $0.3197 \pm 0.0290$ | $0.2238 \pm 0.0112$ | $0.3606 \pm 0.0201$ |

Table 4: Effects of mutations on meta-metrics given 1000 data. CKL values scaled by $P(X_i = \oslash)$ for direct comparison to KL.

Table 2 also shows edit distances and $WED_D$ from the true model. Table 3 and 4 show results for KL and our three $CKL$ meta-metrics with perfect (infinite) and imperfect (1000 cases) datasets respectively. CKL results have been scaled by $\frac{1}{P(X_i=\oslash)}$ to make results directly comparable with KL.

1. **Sensitivity:** Our KL based meta-metrics are all sensitive to strong effects. In each case the "strong" mutation has a larger effect than the "weak" mutation. When $M_1$ and $M_2$ do not contain any reversed arcs (i.e., tweak, add and del), $CKL_3$ and $KL$ are identical. $CKL_2$ is similar and $CKL_1$ is a bit less so. This is expected as a result of $CKL_3$ satisfying Desideratum 1a.

For this example, $WED_D$ is more sensitive to strong connections than weak connections, while $ED$ is blind to arc strength.

2. **Causality:** All meta-metrics except $KL$ are sensitive to the true causal orientation of arcs. The ED based metrics double count a reversed arc so a reversal is seen as twice as bad as an addition or deletion. In this particular example, $CKL_1$, $CKL_2$ and $CKL_3$ all penalize reversals more severely than additions/deletions. This trend is likely to show up in most networks.

3. **Non-negative scalar:** Clearly, all values in our Tables 2, 3 and 4 are non-negative scalars.

4. **Uniqueness:** $ED$ and $WED_D$ fail the Uniqueness Desideratum as they give a perfect score when $M_1 = M_2$ irrespective of $\theta_1$ and $\theta_2$. $KL$ fails as models within the true pattern are all given a score of 0. The $CKL$ only give perfect scores to the true model, or the true model with strictly spurious arcs. The models "add.weak" and "add.strong" have strictly spurious arcs when they are parameterized with infinite data, as in Table 3. The same arcs in Table 4 are parameterized by finite data, and as such are not strictly spurious, they cause a small change in the JPD. This is why these models have a worse score then the true model for finite data, but an identical score for infinite data.

5. **Consistency:** All meta-metrics give a score of 0 to the true model when parameterized from infinite data. As such, all appear to be consistent on this example.

## 6. CONCLUSION

We have noted the need for a more refined means of assessing the learning of causal discovery algorithms than simply eyeballing the networks produced, or its algorithmic correlate of computing edit distance. Kullback Leibler divergence is sensitive to the variations in parameterization required, however it is insensitive to many of the variations in causal structure which can distinguish between good and bad causal discovery. By extending Kullback Leibler to the augmentation (intervention) space in $CKL$ (or Parentset Weighted Edit Distance) we have found a means of overcoming both failings.

## Appendix A. Notation

$\oslash$: Non-intervention state

$M_t$: True model structure

$M_1, \ldots, M_n$: Model structures

$M'_i$: An augmented model structure

$M_i(\theta_i)$: Parameterized model

**X**: Set of variables $\{X_1, \ldots X_n\}$

**X'**: Set of augmented variables such that $\mathbf{X} \cap \mathbf{X'} = \emptyset$

$x_i$: An instance of $X_i$ such that $X_i = x_i$

$\vec{x}_i$ : Vector of values $< x_1, \ldots, x_n >$, an instantiation of **X**

$P_i(x)$ Probability of $X = x$ in parameterized model $M_i(\theta_i)$

$\pi_j(X_i)$: Parents of $X_i$ in model $M_j$

$\pi'_j(X_i)$: Parents of $X_i$ in augmented model $M'_j$

$\vec{y}_{ji}$: Vector of values $< y_1, \ldots, y_n >$ such that $\mathbf{Y} = \pi_j(X_i)$

## Appendix B. Development of $CKL_3$

We prove that there is a reduced class of CKL variants which satisfy all our desiderata, and find the single variant which maximized causal distinguishability.

In its most general form we define CKL as:

$$\begin{aligned}
CKL &= CKL(M_1(\theta_1), M_2(\theta_2)) \\
&= KL(M'_1(\theta'_1), M'_2(\theta'_2)) \\
&= \sum_{\vec{x}} \sum_{\vec{x}'} P'_1(\vec{x}') P'_1(\vec{x}|\vec{x}') \ln \frac{P'_1(\vec{x}') P'_1(\vec{x}|\vec{x}')}{P'_2(\vec{x}') P'_2(\vec{x}|\vec{x}')} \\
&= \sum_{\vec{x}'} P'_1(\vec{x}') \sum_{\vec{x}} P'_1(\vec{x}|\vec{x}') \ln \frac{P'_1(\vec{x}|\vec{x}')}{P'_2(\vec{x}|\vec{x}')} \\
&\quad + \sum_{\vec{x}'} P'_1(\vec{x}') \ln \frac{P'_1(\vec{x}')}{P'_2(\vec{x}')}
\end{aligned}$$

where $P'_1(\vec{x}')$ and $P'_2(\vec{x}')$ are arbitrary probability distributions.

To simplify, the following assumptions are made:

1. There is a one-to-one mapping of nodes from $X_i$ to $X'_i$. Each node in **X** has a corresponding intervention variable in **X'**

2. $\pi'_j(X_i) = \pi_j(X_i) \cup \{X'_i\}$

   The augmented parent set of $X_i$ (i.e., $\pi'(X_i)$) is the the unaugmented parent set of $X_i$ (i.e., $\pi(X_i)$) with the added parent $X'_i$.

3. We limit our intervention variables to taking $|X_i|+1$ values, namely each value which $X_i$ can take plus $\oslash$, the "no-intervention" state. Each intervention node $X`_i$ has total control over its counterpart $X_i$: it may set $X_i$ to any of its values, or not intervene at all.

4. $P'_j(x_i|\pi'_1(x_i)) = \begin{cases} P_j(x_i|\pi_1(x_i)) & \text{if } x'_i = \oslash \text{ (i.e., no intervention)} \\ 1.0 & \text{if } x'_i = x_i \text{ (i.e., intervened as } x'_i) \\ 0.0 & \text{otherwise} \end{cases}$

   For simplicity, we limit our interventions to perfect interventions (interventions which cannot fail).[17] If no intervention is performed, the original probability distribution is retained.

These assumptions do not limit the choice of distributions over $P'(x')$ in any way. An arbitrary distribution may be represented as being "passed through" the intervention nodes. All indirect intervention variables can then be combined into an equivalent distribution over $P'(\vec{x}')$. In this way the network structure is restricted but possible intervention distributions remain unrestricted.

Using this restricted class of network structure:

$$
\begin{aligned}
CKL \;=\; & \sum_{\vec{x}'}^{\prod(\mathbf{X}')} P'_1(\vec{x}') \sum_{i=0}^{|\mathbf{X}|} \sum_{\vec{y}'_{1i}}^{\prod(\pi'_1(X_i))} P'_1(\vec{y}'_{1i}|\vec{x}') \sum_{x_i \in X_i} P'_1(x_i|\vec{y}'_{1i}, \vec{x}') \ln P'_1(x_i|\vec{y}'_{1i}, \vec{x}') \\
& - \sum_{\vec{x}'}^{\prod(\mathbf{X}')} P'_1(\vec{x}') \sum_{i=0}^{|\mathbf{X}|} \sum_{\vec{y}'_{2i}}^{\prod(\pi'_2(X_i))} P'_1(\vec{y}'_{2i}|\vec{x}') \sum_{x_i \in X_i} P'_1(x_i|\vec{y}'_{2i}, \vec{x}') \ln P'_2(x_i|\vec{y}'_{2i}, \vec{x}') \\
& + \sum_{\vec{x}'}^{\prod(\mathbf{X}')} P'_1(\vec{x}') \ln \frac{P'_1(\vec{x}')}{P'_2(\vec{x}')}
\end{aligned}
$$

As we use perfect interventions, cases where $x'_i \neq \oslash$ do not contribute to the sum as $1\ln(1) = 0$ and by convention $0\ln(0) = 0$. Using this we replace $P'$ with $P$ as per assumption 4. Summands are rearranged for clarity.

---

17. Imperfect interventions also work under this framework. An imperfect intervention can be modelled as a two-stage process: an imperfect "attempted intervention", which causes a perfect intervention when it succeeds, or a non-intervention when it fails. A similar argument holds for interventions which impact on a third variable. Any desired set of imperfect interventions may be mapped onto a distribution over the set of perfect interventions $\mathbf{X}'$.

$$CKL = \sum_{i=0}^{|\mathbf{X}|} \sum_{\vec{y}'_{1i}}^{\prod(\pi'_1(X_i))} \sum_{\vec{x}'}^{\prod(\mathbf{X}')} P'_1(\vec{x}', \vec{y}'_{1i}, x'_i = \oslash) \sum_{x_i \in X_i} P_1(x_i|\vec{y}_{1i}) \ln P_1(x_i|\vec{y}_{1i}) \qquad (2)$$

$$- \sum_{i=0}^{|\mathbf{X}|} \sum_{\vec{y}'_{2i}}^{\prod(\pi'_2(X_i))} \sum_{\vec{x}'}^{\prod(\mathbf{X}')} P'_1(\vec{x}', \vec{y}'_{2i}, x'_i = \oslash) \sum_{x_i \in X_i} P'_1(x_i|\vec{y}'_{2i}, \vec{x}') \ln P_2(x_i|\vec{y}_{2i})$$

$$+ \sum_{\vec{x}'}^{\prod(\mathbf{X}')} P'_1(\vec{x}') \ln \frac{P'_1(\vec{x}')}{P'_2(\vec{x}')}$$

### B.1 CKL vs KL when $M_1 = M_2$

Equation 2 shows CKL in its most general form, however as our goal here is to find what augmentations, $P'_1(\mathbf{X}')$ and $P'_2(\mathbf{X}')$, set $KL = CKL \times c + d$ when $M_1$ and $M_2$ share a common DAG:

We define $CKL_O(M_1(\theta_1), M_2(\theta_2))$ as $CKL(M_1(\theta_1), M_2(\theta_2))$ where $M_1 = M_2$. $KL_O$ is defined in the same manner. As $CKL_O$ network structures are identical $\pi_1(X_i) = \pi_2(X_i)$ and $\pi'_1(X_i) = \pi'_2(X_i)$, as such the $\pi_j$ subscript is omitted to reduce clutter.

With this in mind:

$$CKL_O = CKL(M_1(\theta_1), M_2(\theta_2)) \text{ where } M_1 = M_2$$

$$= \sum_{i=0}^{|\mathbf{X}|} \sum_{\vec{y}'_i}^{\prod(\pi'(X_i))} \sum_{\vec{x}'}^{\prod(\mathbf{X}')} P'_1(\vec{x}', \vec{y}'_i, x'_i = \oslash) \sum_{x_i}^{\prod(X_i)} P_1(x_i|\vec{y}_i) \ln P_1(x_i|\vec{y}_i)$$

$$- \sum_{i=0}^{|\mathbf{X}|} \sum_{\vec{y}'_i}^{\prod(\pi'(X_i))} \sum_{\vec{x}'}^{\prod(\mathbf{X}')} P'_1(\vec{x}', \vec{y}'_i, x'_i = \oslash) \sum_{x_i}^{\prod(X_i)} P_1(x_i|\vec{y}_i) \ln P_2(x_i|\vec{y}_i)$$

$$+ \sum_{\vec{x}'}^{\prod(\mathbf{X}')} P'_1(\vec{x}') \ln \frac{P'_1(\vec{x}')}{P'_2(\vec{x}')}$$

Replacing $\pi'_i$ with $\pi_i \cup \{X'_i\}$ and simplifying:

$$CKL_O = \sum_{i=0}^{|\mathbf{X}|} \sum_{\vec{y}_i}^{\prod(\pi(X_i))} \sum_{\vec{x}'}^{\prod(\mathbf{X}')} P'_1(\vec{x}', \vec{y}_i, x'_i = \oslash) \sum_{x_i}^{\prod(X_i)} P_1(x_i|\vec{y}_i) \ln \frac{P_1(x_i|\vec{y}_i)}{P_2(x_i|\vec{y}_i)} \qquad (3)$$

$$+ \sum_{\vec{x}'}^{\prod(\mathbf{X}')} P'_1(\vec{x}') \ln \frac{P'_1(\vec{x}')}{P'_2(\vec{x}')}$$

Applying the same logic to KL:

$$KL_O(M_1(\theta_1), M_2(\theta_2)) = \sum_{i=0}^{|\mathbf{X}|} \sum_{\vec{y}_i}^{\prod(\pi(X_i))} P_1(\vec{y}_i) \sum_{x_i}^{\prod(X_i)} P_1(x_i|\vec{y}_i) \ln \frac{P_1(x_i|\vec{y}_i)}{P_2(x_i|\vec{y}_i)} \qquad (4)$$

Next, we need to quantify $P'_1$ and $P'_2$ in Equation 3 satisfying:

$$KL_O(M_1(\theta_1), M_2(\theta_2)) = CKL_O(M_1(\theta_1), M_2(\theta_2)) \times c + d$$

However, for any choice of $P'_1(\vec{x}')$ such that $CKL_O \leq KL_O$ a corresponding function $P'_2(\vec{x}')$ may be chosen to increase the value of $CKL_O$.[18] Likewise $CKL_O$ can be artificially reduced by changing the frequency of the $P'_1(X = \text{"All intervention"})$ state.

It seems reasonable to make the further restriction that $CKL_O$ and $KL_O$ should still be equivalent given evidence $\vec{e}$. It may be useful to assess how well a network performs in a subdomain, for example when smoking is always $True$, or gender is always $Male$. KL and CKL can both be used to assess subdomains by replacing $P(\vec{x}|\vec{y})$ with $P(\vec{x}|\vec{y},\vec{e})$.

So

$$KL_0(\mathbf{X}|\vec{e}) = CKL_0(\mathbf{X}|\vec{e}) \times c + d$$

$$\sum_{\vec{x}}^{\prod(\mathbf{X})} P_1(\vec{x}|\vec{e}) \ln \frac{P_1(\vec{x}|\vec{e})}{P_2(\vec{x}|\vec{e})} = \sum_{\vec{x}}^{\prod(\mathbf{X})} \sum_{\vec{x}'}^{\prod(\mathbf{X}')} P'_1(\vec{x}'|\vec{e}) P'_1(\vec{x}|\vec{x}',\vec{e}) \ln \frac{P'_1(\vec{x}|\vec{x}',\vec{e})}{P'_2(\vec{x}|\vec{x}',\vec{e})} \times c$$

$$+ \sum_{\vec{x}'}^{\prod(\mathbf{X}')} P'_1(\vec{x}'|\vec{e}) \ln \frac{P'_1(\vec{x}'|\vec{e})}{P'_2(\vec{x}'|\vec{e})} \times c + d$$

where $\vec{e}$ is some set of evidence from $\mathbf{X}$.

Following from this, it is clear that, since $KL_O(\mathbf{X}|\vec{e}) = CKL_O(\mathbf{X}|\vec{e}) = 0$ when $\vec{e} \in \prod(\mathbf{X})$:

$$\sum_{\vec{x}'}^{\prod(\mathbf{X}')} P'_1(\vec{x}'|\vec{e}) \ln \frac{P'_1(\vec{x}'|\vec{e})}{P'_2(\vec{x}'|\vec{e})} = 0$$

$$\therefore P'_1(\vec{x}'|\vec{e}) = P'_2(\vec{x}'|\vec{e})$$

$$\therefore d = 0$$

That is, the $KL$ and $CKL$ between two fully observed networks is 0, so, to keep our original equality, the second term of CKL is necessarily 0. This result can also be obtained from Desideratum 4 which requires the true model to have a score of 0. This results also implies $d = 0$.

With this in mind:

$$CKL_O = \sum_{i=0}^{|\mathbf{X}|} \sum_{\vec{y}_i}^{\prod(\pi(X_i))} \sum_{\vec{x}'}^{\prod(\mathbf{X}')} P'_1(\vec{x}', \vec{y}_i, x'_i = \oslash) \sum_{x_i}^{\prod(X_i)} P_1(x_i|\vec{y}_i) \ln \frac{P_1(x_i|\vec{y}_i)}{P_2(x_i|\vec{y}_i)} \qquad (5)$$

We can see from equation 4 and 5 that $KL_O = CKL_O \times c$ when:

$$\forall i \in |\mathbf{X}|, \vec{y}_i \in \prod(\pi(X_i)): \ P_1(\vec{y}_i) = \sum_{\vec{x}'}^{\prod(\mathbf{X}')} P'_1(\vec{x}', \vec{y}_i, x'_i = \oslash) \times c$$

$$\therefore P_1(\vec{y}_i) = P'_1(\vec{y}_i, x'_i = \oslash) \times c$$

---

18. This trick doesn't work when $CKL_O < KL_O$ as $\forall P'_1(\cdot), P'_2(\cdot): \sum_{\vec{x}'}^{\prod(\mathbf{X}')} P'_1(\vec{x}') \ln \frac{P'_1(\vec{x}')}{P'_2(\vec{x}')} \geq 0$

It can also be shown that $P'_1(x'_i = \oslash)$ must be a constant.

$$\forall i \in |\mathbf{X}| : \sum_{\vec{y}_i}^{\prod(\pi(X_i))} P_1(\vec{y}_i) = \sum_{\vec{y}_i}^{\prod(\pi(X_i))} P'_1(\vec{y}_i, x'_i = \oslash) \times c$$
$$\forall i \in |\mathbf{X}| : 1 = P'_1(x'_i = \oslash) \times c$$
$$\therefore P'_1(x'_i = \oslash) = \frac{1}{c}$$
$$\therefore P_1(\vec{y}_i) = P'_1(\vec{y}_i | x'_i = \oslash)$$

Summarizing, in our ideal CKL metric, the joint distribution over the parents of each node in $P$ is equal to the joint distribution over the parents of each node in $P'$ when that node is NOT being intervened upon. This comes as a direct result of Desideratum 1a; we refer to this as the PJP (Parent Joint Probability) condition below.

### B.2 Maximal Causal Discrimination

Given an intervention set, the distribution over intervention values is given above by the PJP condition. However, the problem of choosing an intervention set has not been addressed.

We have shown that there are some constraints on the distribution over intervention sets, namely $P'_1(x'_i = \oslash)$ is constant for all $i$. It can also be shown that many intervention sets are invalid.

Consider the network in Figure 9. If our intervention sets contains Pregnant but not Gender or Hormone, there is no distribution $P'(\pi(X_i))$ satisfying the PJP condition when $X_i = Hormone$. This is clearly the case as an intervention on Pregnant is not affected by the observation of Gender, so in $P'$ pregnant males are possible, while in $P$ they are not. This argument also holds for less extreme distributions.

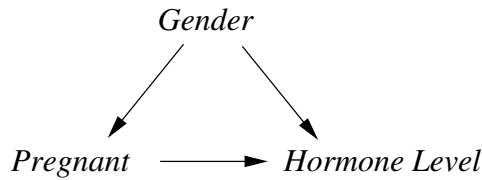

Figure 9: Network showing relationship between Gender, Pregnancy and some Hormone effected by both.

Clearly there are some distributions where this PJP condition holds. It is trivially true for the "All non-intervention" case, where CKL is equivalent to KL. At the opposite extreme, it is also true for the "All-but-one" intervention case where one node at a time from $\mathbf{X}'$ is set to $\oslash$, with the intervention distribution over all other variables are set as per $P$.

Between these two extremes are shades of grey in which a subset of variables may be intervened upon. By inspection, it appears that, at least for fully connected networks,[19] the PJP condition is possible when our non-intervention nodes fall into a contiguous subnetwork. That is the set of non-intervention variables must form a contiguous block in the total order of $M_1$. In Figure 9 the subnetwork of non-intervention variables, $\{Gender, Hormone\}$ is not contiguous.

It is possible to set $P'(\mathbf{X}')$ as a mixture of every distribution satisfying the PJP condition, so long as $P'_1(x'_i = \oslash)$ remains uniform. Any such distribution will fulfill Desideratum 1a. However, as mentioned previously, setting $P'_1(x'_i = \oslash) = 1.0$ will fulfill this criterion and is simply the KL distance we were originally attempting to improve.

To avoid this, a distribution which maximally discriminates between causally distinct models is suggested. Any choice of intervention set which includes multiple non-intervention values may fail to causally discriminate between those values. The logical thing to do is use an intervention set distribution which is uniform over all intervention sets containing a single non-intervention value.

This gives the single distribution which both optimally scores probability distributions when the true causal structure is known, and maximally penalizes causal discrepancies in the network structure.

---

19. Any non-fully connected network may be represented by a fully connected network with zero weight arcs.